\documentclass{article}

\usepackage{subcaption}

\usepackage{arxiv}

\usepackage[utf8]{inputenc} 
\usepackage[T1]{fontenc}    

\usepackage{url}            
\usepackage{booktabs}       
\usepackage{amsfonts}       
\usepackage{nicefrac}       
\usepackage{microtype}      
\usepackage{graphicx}
\usepackage[square,numbers]{natbib}
\usepackage{wrapfig}

\usepackage{doi}

\usepackage{amsmath}
\usepackage{tabularx}
\usepackage{multicol}
\usepackage{multirow}
\usepackage{caption}
\usepackage{hhline}
\usepackage{bbding}
\usepackage{makecell}
\usepackage{array}
\usepackage{bm}
\usepackage{relsize}

\usepackage{adjustbox}

\usepackage[accsupp]{axessibility}  
\usepackage{orcidlink}

\newcolumntype{?}{!{\vrule width 1pt}}

\newcommand\freefootnote[1]{%
  \let\thefootnote\relax%
  \footnotetext{#1}%
  \let\thefootnote\svthefootnote%
}

\author{Ali Abbasi\\
Vanderbilt University\\
{\tt\small ali.abbasi@vanderbilt.edu}
\And
Ashkan Shahbazi\\
Vanderbilt University\\
{\tt\small ashkan.shahbazi@vanderbilt.edu}
\And
Hamed Pirsiavash\\
University of California, Davis\\
{\tt\small hpirsiav@ucdavis.edu}
\And
Soheil Kolouri\\
Vanderbilt University\\
{\tt\small soheil.kolouri@vanderbilt.edu}
}

\title{One Category One Prompt: \\ Dataset Distillation using Diffusion Models} 
\shorttitle{One Category One Prompt: Dataset Distillation using Diffusion Models}

\date{}

\hypersetup{
pdftitle={Condensing ImageNet into a Thousand Prompts: Dataset Distillation Using Foundation Models},
pdfauthor={Ali Abbasi, Ashkan Shahbazi, Hamed Pirsiavash, Soheil Kolouri},
pdfkeywords={Dataset Distillation, Diffusion Generative Models},
}

\begin{document}
\maketitle
\vspace{-0.2in}
\begin{abstract}
The extensive amounts of data required for training deep neural networks pose significant challenges on storage and transmission fronts. Dataset distillation has emerged as a promising technique to condense the information of massive datasets into a much smaller yet representative set of synthetic samples. However, traditional dataset distillation approaches often struggle to scale effectively with high-resolution images and more complex architectures due to the limitations in bi-level optimization. Recently, several works have proposed exploiting knowledge distillation with decoupled optimization schemes to scale up dataset distillation. Although these methods effectively address the scalability issue, they rely on extensive image augmentations requiring the storage of soft labels for augmented images. In this paper, we introduce Dataset Distillation using Diffusion Models (D3M) as a novel paradigm for dataset distillation, leveraging recent advancements in generative text-to-image foundation models.  Our approach utilizes textual inversion, a technique for fine-tuning text-to-image generative models, to create concise and informative representations for large datasets. By employing these learned text prompts, we can efficiently store and infer new samples for introducing data variability within a fixed memory budget. We show the effectiveness of our method through extensive experiments across various computer vision benchmark datasets with different memory budgets. 
\end{abstract}

\freefootnote{Preprint.}

\section{Introduction}
\label{sec:intro}

The substantial data needed for training deep neural networks \cite{OpenImages2} imposes significant burdens on storage, transmission, and handling, impeding tasks that necessitate repeated training on these datasets, like hyperparameter optimization \cite{maclaurin2015gradient,lorraine2020optimizing}. Moreover, the publication of such extensive datasets gives rise to copyright and privacy concerns, further complicating their utilization. These challenges have prompted an important scientific question first posed in \cite{wang2018dataset}: How much data is encoded in a given training set? Can one construct a small subset of synthetic samples such that models trained on this subset achieve competitive performance compared to training on the entire dataset? This problem, referred to as Dataset Distillation \cite{wang2018dataset} or Dataset Condensation \cite{zhao2020dataset}, has since spurred a notable body of research from the community \cite{liu2023slimmable,cazenavette2023generalizing,cui2023scaling,sun2023diversity,guo2023towards,shao2023generalized,zhou2023dataset,yin2024squeeze,chen2024data}.

Dataset distillation is often properly formalized as a bi-level optimization problem, where the inner optimization focuses on training a model on the distilled set (i.e., a small set of synthetic samples), while the outer (meta) optimization focuses on refining the distilled set \cite{wang2018dataset} to improve the result of the inner optimization. Such bi-level optimization presents significant challenges, as evaluating the outer optimization loop necessitates solving the inner optimization loop and thus requires backpropagation of errors through the entire inner training process, which is memory-intensive and computationally expensive. Many research papers are devoted to devising ways to ameliorate the challenges in this bi-level optimization. For instance, by introducing surrogate objectives for computing the meta gradients through gradient matching \cite{zhao2020dataset} or training trajectory matching \cite{cazenavette2022dataset}, among others. Despite substantial efforts in this area, existing methods employing bi-level optimization often struggle to scale up to larger datasets and models \cite{yin2024squeeze}. To scale up dataset distillation to ImageNet-1K  \cite{deng2009imagenet}, recent works focus on decoupling the bi-level optimization into two single-level learning procedures \cite{yin2024squeeze,sun2023diversity}.

\begin{figure}[t!]
    \centering
    \includegraphics[width=\linewidth]{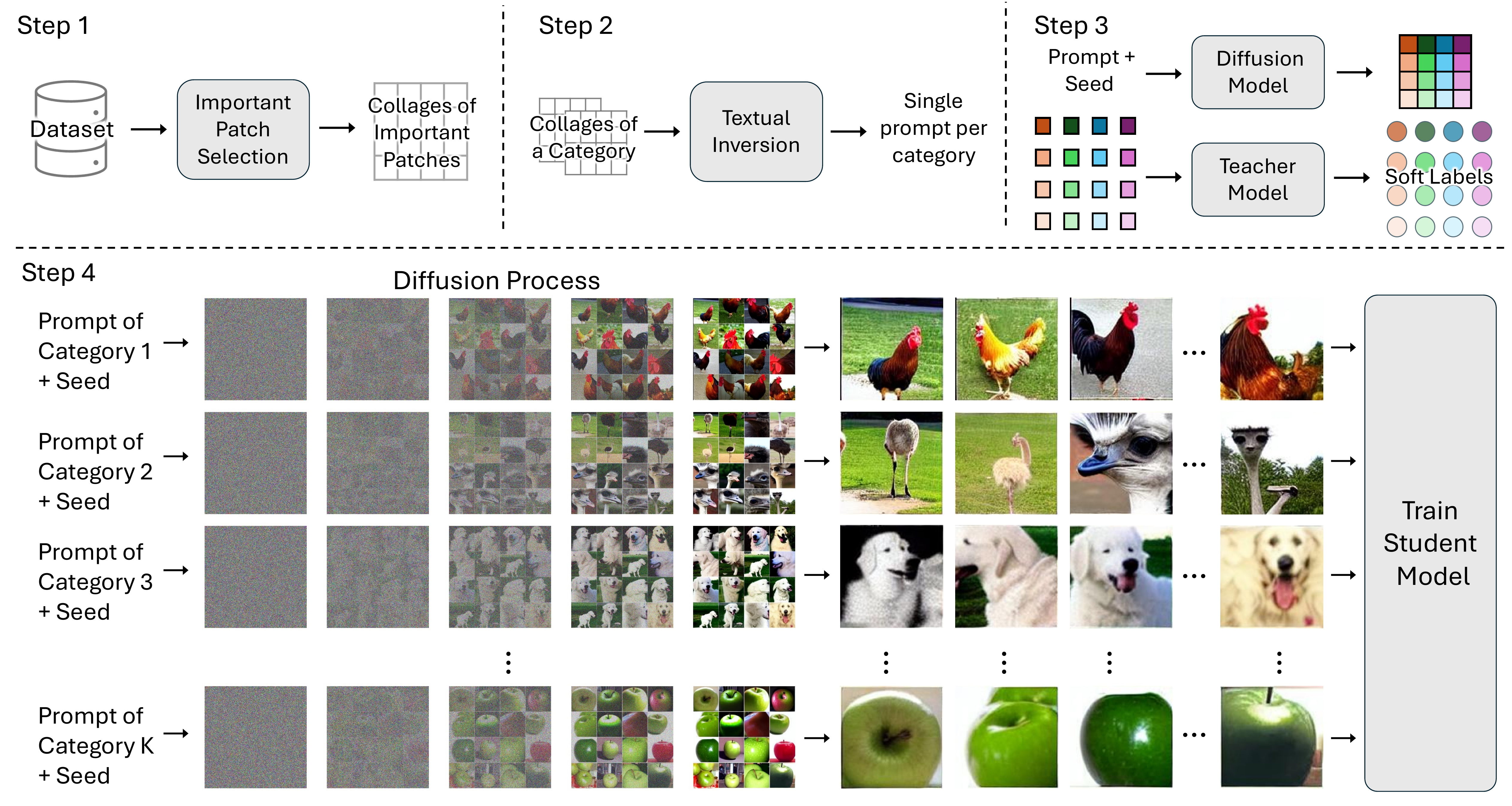}
    \caption{Illustrating the core steps in our proposed framework, Dataset Distillation using Diffusion Models (D3M). Step 1 follows the work of \cite{sun2023diversity} and utilizes a teacher network to identify important patches of the training data and create collages of these patches. Step 2 employs textual inversion \cite{gal2022image} to optimize a single prompt per category, resulting in the creation of collage images through stable diffusion \cite{rombach2022high}. Regarding labels, we consider two different settings, namely, one-hot and soft-labels. To generate soft-labels for synthetic images of each category, in Step 4, a random seed is fixed, and stable diffusion is utilized to generate collage images, which are then fed to the teacher network to obtain the soft-labels. Finally, in Step 4, the categorical prompts and random seeds are employed to create the distilled dataset and train the student.}
    \vspace{-.3in}
    \label{fig:teaser} 
\end{figure}
\vspace{-.3cm}
Another significant challenge in dataset distillation is that the distilled synthetic dataset is often optimized with respect to a specific network architecture, limiting its ability to generalize well to other architectures \cite{cazenavette2023generalizing}. On this front, generating realistic images has emerged as a powerful method to improve the performance and generalizability of dataset distillation methods. In short, generating synthetic samples closer to the training data manifold has been shown to enhance the generalizability of the distilled datasets across diverse architectures \cite{cazenavette2023generalizing,yin2024squeeze,sun2023diversity}. This can be achieved, for instance, by utilizing generative models \cite{cazenavette2023generalizing}, batch statistics of pretrained models on the training dataset \cite{yin2024squeeze}, or simply by creating synthetic samples through collating patches from the training data \cite{sun2023diversity}.


In this paper, inspired by recent advances in foundation models \cite{brown2020language,radford2021learning,rombach2022high}, and following the original scientific curiosity of Wang et al. \cite{wang2018dataset}, we pose the following question: \textbf{how compressible is a dataset conditioned on having access to a foundation model?}  This is an important scientific inquiry as such compression, if possible, can significantly reduce communication and storage costs. Moreover, it is not far-fetched to consider a foundation model as a universal data converter that exists on all clients or is accessible via an API. \textbf{In this work, we demonstrate that an entire category of ImageNet can be condensed into a single prompt of latent diffusion models \cite{rombach2022high}}, leading to state-of-the-art dataset distillation performance. 
Our proposed framework, denoted as Dataset Distillation using Diffusion Models (D3M), builds on recent advances in dataset distillation \cite{cazenavette2023generalizing,yin2024squeeze,sun2023diversity}, latent diffusion models \cite{rombach2022high}, and textual inversion \cite{gal2022image}, enabling unprecedented compression of ImageNet-scale datasets while providing competitive training performance and cross-architecture generalization.

Our specific \textbf{contributions} in this paper are:
\begin{enumerate}
    \item Demonstrating the potential of diffusion models \cite{rombach2022high} for dataset distillation, showing unprecedented condensation rates.
    \item Adapting textual inversion \cite{gal2022image} to dataset distillation, and demonstrating the possibility of generating realistic and diverse collages of images representing an image category via a single prompt. 
    \item To the best of our knowledge, among large-scale dataset distillation methods, we are the first to offer a solution for the memory overhead associated with storing soft labels for augmentations and to demonstrate its trade-offs. 
    
\end{enumerate}

\section{Related Work}
The current research on dataset distillation can be broadly categorized into the following groups: 1) those that formulate the problem as bi-level optimization, 2) those that simplify the bi-level optimization problem into a uni-level optimization, 3) core-set methods, and 4) methods that utilize generative or implicit priors for improved alignment with the data manifold. Below, we briefly discuss these categories and the corresponding papers.

\textbf{Bi-level optimization-based dataset distillation.} Dataset distillation can be conceptualized as a bi-level meta-learning problem \cite{wang2018dataset}, where the outer loop is responsible for optimizing the distilled dataset, while the inner loop focuses on training a model using this distilled dataset. To address the computational and memory complexities associated with bi-level optimization, existing literature focuses on devising surrogate objectives for computing the outer-level gradients. For example, \cite{zhao2020dataset,kim2022dataset,zhang2023accelerating} utilize gradient matching for the outer-level optimization, \cite{wang2022cafe,zhao2023dataset} employ feature and distribution alignment, and \cite{cazenavette2022dataset,cui2022dc,guo2023towards} leverage training trajectory matching/alignment. Particularly, trajectory alignment approaches have shown outstanding performance for dataset distillation on small-scale datasets, such as CIFAR-10. However, bi-level optimization methods encounter two major challenges: 1) scaling up to higher-resolution datasets and larger models, and 2) generalization to diverse architectures. We do not do bi-level optimization so we can scale up to large datasets and generalize better.

\textbf{Uni-level relaxation of dataset distillation.} 
 A theoretically appealing line of research focuses on Neural Tangent Kernels (NTKs) \cite{jacot2018neural,novak2019neural}, which offer a closed-form solution (i.e., the solution to Kernel Ridge Regression) for the inner optimization problem (assuming infinitely wide neural networks), effectively converting the bi-level optimization problem into a uni-level optimization \cite{nguyen2021dataset,loo2023dataset}. These methods have demonstrated remarkable efficacy in dataset distillation with small-scale datasets, bolstered by robust theoretical underpinnings. Unfortunately, they face limitations in scaling to datasets featuring higher-resolution images and larger models. An alternative approach \cite{yin2024squeeze, sun2023diversity} aims to address this challenge by breaking down the bi-level optimization problem into two uni-level, decoupled optimization problems. While relinquishing claims to optimality, these decoupled optimizations have proven effective in scaling dataset distillation to ImageNet-scale datasets and larger networks (e.g., ResNets). Inspired by this line of research, we also leverage a uni-level optimization approach in this paper.

\textbf{Coreset selection for efficient machine learning.} Unlike classic dataset distillation frameworks \cite{wang2018dataset}, which concentrate on generating a small set of synthetic samples, coreset selection methods \cite{mirzasoleiman2020coresets,pooladzandi2022adaptive,maalouf2022unified} prioritize identifying a small subset of the training set that enables training a model with competitive performance compared to training on the entire dataset. A potential advantage of coreset selection methods is that, by definition, the coreset belongs to the data manifold, thereby offering superior generalization across architectures. Notably, many coreset methods also utilize a bi-level optimization to find the core subset \cite{borsos2020coresets}. Other coreset approaches focus on devising difficulty-based metrics to assess the sample importance, e.g., the forgetting \cite{toneva2018empirical} and the EL2N scores \cite{paul2021deep}. The images in our distilled dataset are not real but look natural, thanks to the remarkable capability of diffusion models.

\textbf{Generative priors for dataset distillation.} Cazenavette et al. \cite{cazenavette2023generalizing} demonstrate the importance of utilizing generative priors for dataset distillation. They show that solving the bi-level optimization problem in the latent space of a generative model to generate the distilled dataset, would enhance the performance of many dataset distillation techniques \cite{cazenavette2022dataset,zhao2023dataset}. A critical observation in \cite{cazenavette2023generalizing}, however, is that a proper amount of generative prior is necessary for cross-architecture generalization, while too strong a prior limits the expressivity and thus hurts distillation performance. Yin et al. \cite{yin2024squeeze} utilize the batch statistics of a teacher model to guide the distilled dataset towards more realistic images, similar to techniques developed in model inversion attacks \cite{fredrikson2015model,yin2020dreaming}. In a different approach, Sun et al. \cite{sun2023diversity} posit that the \emph{diversity} and \emph{realism} of synthetic samples are critical for dataset distillation methods. Moreover, they observe that a large portion of input samples, e.g., the background in images, do not contain valuable information for the downstream task, and hence suggest creating diverse collages of important patches (i.e., foregrounds) addressing diversity and realism. Interestingly, the generated collages are qualitatively similar to the distilled images in \cite{cazenavette2023generalizing} for the ``proper amount of generative prior.'' Inspired by these works, our method utilizes latent diffusion models \cite{rombach2022high} together with textual inversion \cite{gal2022image} to generate diverse distilled samples that focus on the foreground of a category. 

\section{Method}
We strive to distill large-scale datasets into condensed representations that maintain high accuracy when employed for training classification or regression models. In line with recent advancements in large-scale data distillation \cite{yin2024squeeze,sun2023diversity}, we steer clear of bi-level optimization. The pivotal components of our proposed framework are outlined as follows:
\begin{enumerate}
    \item Prioritizing informative patches from training images and optimizing the utilization of the number of images per category by generating collage images composed of these important patches, akin to the approach outlined in \cite{sun2023diversity}.
    \item Instead of directly storing collages of important patches for each category, we utilize a text-to-image diffusion model and employ textual inversion techniques \cite{gal2022image} to generate prompts that directly create the collage images, enabling the model to produce desired collages on demand.
\end{enumerate}
During classifier training, the diffusion model can be applied to the stored prompts using either in-house or API-based services. This allows for the efficient generation of collage images, which can then be effectively utilized for classifier training. Our approach is motivated by the enhanced dataset condensation enabled by the low dimensionality of textual prompts in text-to-image diffusion models. Additionally, we leverage the remarkable capability of generative models to concentrate the most discriminative details of an entire category using just a single low-dimensional prompt vector (e.g., 768 scalars).


Our proposed framework, denoted as Dataset Distillation using Diffusion Models (D3M), is shown in Figure \ref{fig:teaser}, and it consists of three main steps for condensing the dataset, and a fourth step for training a classifier/regressor. Below, we delve into a detailed explanation of these steps.

\begin{figure}[t!]
    \centering
    \includegraphics[width=\linewidth]{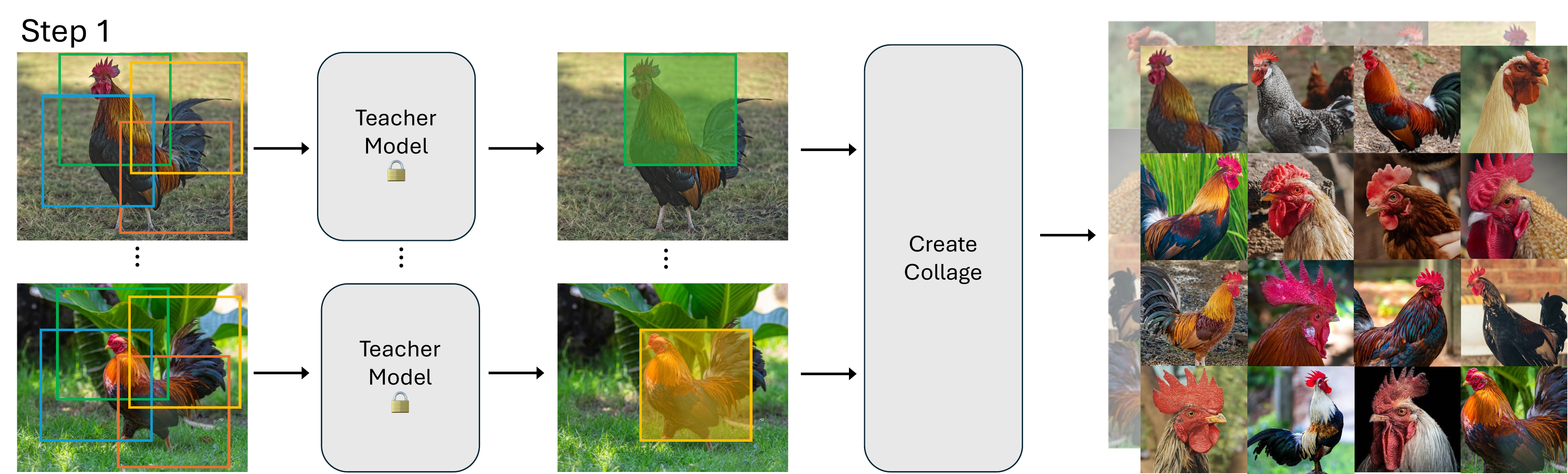}
    \caption{Following the work of Sun et al. \cite{sun2023diversity}, we first identify a patch per input image that results in the lowest cross-entropy loss for a pre-trained and frozen teacher model. Then, we construct collage images of these important patches.}
    \label{fig:step1}
\end{figure}

\subsection{Step 1: Collage Generation with Important Patches}

Let us denote the training data as $\mathcal{D}:=\{(x_i,y_i)\}_{i=1}^N$, where $x_i\in \mathbb{R}^{H\times W \times 3}$ represents the $i$'th training image, and $y_i\in \mathbb{R}^K$ is the corresponding label. Moreover, we denote the data belonging to category $c$ as $\mathcal{D}_c:=\{x_i|y_i=c\}_{i=1}^{N_c}$. Additionally, let $f:\mathbb{R}^{H\times W \times 3}\to \mathbb{R}^K$ denote a pretrained teacher model on dataset $\mathcal{D}$. Our goal in collage generation is to first identify an informative patch for each image, denoted as $x'_i\in \mathbb{R}^{H_p\times W_p\times 3}$. To achieve this, we follow the methodology of Sun et al. \cite{sun2023diversity} and solve the following optimization problem:
\begin{equation}
x^*_i = \operatorname{argmax}_{x'\sim p(x'_i|x_i)} \ell(f(x'),y_i)
\end{equation}
where $\ell(\cdot,\cdot)$ denotes the cross-entropy loss, and $p(x'_i|x_i)$ is the probability density function of $H_p\times W_p$ patches of the $H\times W$ image. We approximate this optimization problem by first randomly sampling a set of patches from the input image, and then feeding them to the pretrained and frozen teacher model, $f$, and selecting the patch with the minimum cross-entropy loss. Alternatively, visual explainability methods like class activation maps (CAM) \cite{zhou2016learning} and its variations \cite{selvaraju2017grad,chattopadhay2018grad} could be used to identify the important patches and guide the collage generation.

Having found the important patches, $\mathcal{D}^*_c=\{x_i^*|y_i=c\}_{i=1}^{N_c}$ we generate grids of these patches to create collage images. We denote these collage images for each category via $\{X^c_j\in\mathbb{R}^{H\times W\times 3}\}_{j=1}^{L_c}$. Figure \ref{fig:step1} demonstrates this concept, when $H=W=512$ and $H_p=W_p=128$, and $X_j^c$ is a $4\times 4$ collage image of important patches from class "cock." 

\subsection{Step 2: Textual Inversion}

In the second step of our framework, our goal is to reduce the generated collage images into textual prompts for a text-to-image diffusion model, e.g., the Latent Diffusion Model (LDM) \cite{rombach2022high}. We aim to find an optimal prompt (per category), which in turn leads to the generation of realistic-looking collage images for each category. We adopt the textual inversion framework proposed by Gal et al. \cite{gal2022image} to optimize such prompts. Throughout the remainder of this subsection, we present the diffusion equations without distinction between whether diffusion is applied in the raw pixel space or in the latent space of an auto-encoder, as our discussion applies to both settings.

\begin{figure}[t!]
    \centering
    \includegraphics[width=\linewidth]{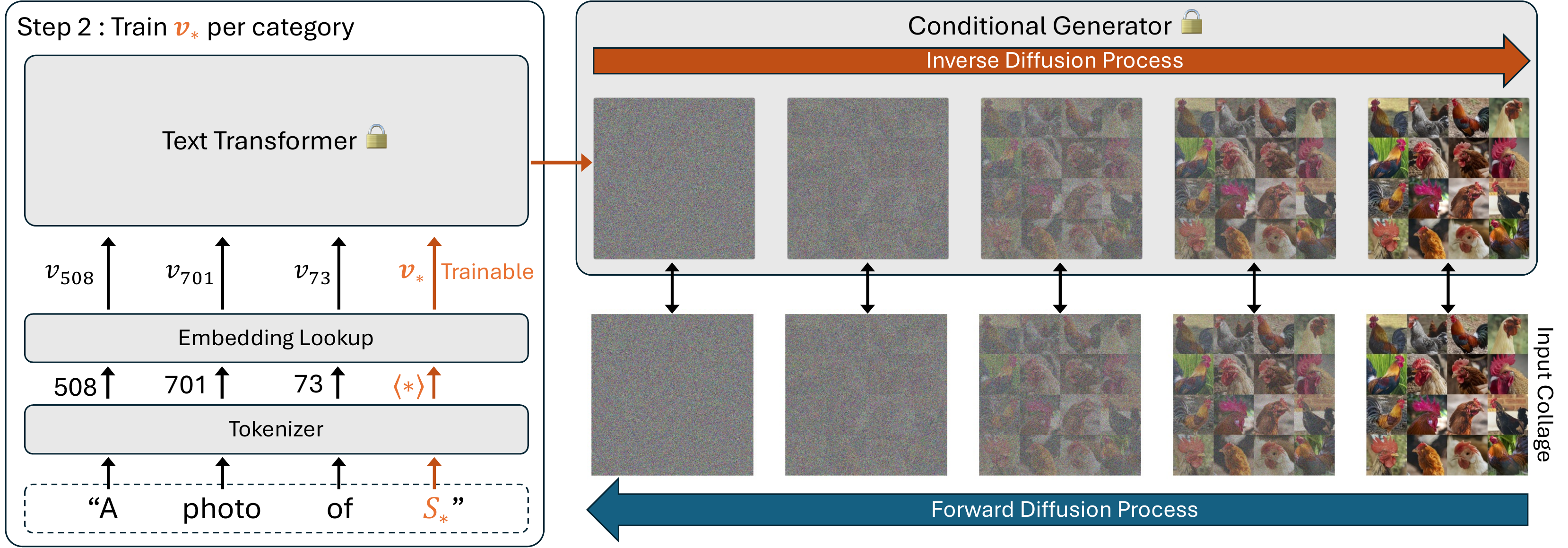}    
    \caption{Given a text-to-image diffusion model, such as Latent Diffusion Model (LDM) \cite{rombach2022high}, for each category of the training data, we employ `textual inversion' \cite{gal2022image} to optimize a token (i.e., a prompt), $v_*$, resulting in the generation of collage images that are similar to the ones constructed in Step 1 (Figure \ref{fig:step1}). }
    \vspace{-.2in}
    \label{fig:step2}
\end{figure}

Let $p_X(c)$ denote the distribution of collage images of category $c$ from Step 1, $X(t)$ denote a collage image $X$ noised to time $t$, and $\epsilon\sim\mathcal{N}(0,1)$ denote an unscaled noise sample used to create $X(t)$. Moreover, let $\phi$ denote the conditional generator (i.e., the denoiser), and $\rho$ denote a text encoder/transformer  (e.g., BERT \cite{devlin2018bert}) that encodes the textual prompt into a conditioning vector for the denoiser.  Following the approach in \cite{gal2022image}, we designate a placeholder string, $S_*$, to represent the new concept we wish to learn, and utilize the following prompt: 
``A photo of $S_*$.'' The word embedding for the placeholder $S_*$ is what we refer to as the "prompt," denoted by $v$. Then, for category $c$, we optimize this prompt via the following optimization problem (see Figure \ref{fig:step2}):
\begin{equation}
v^c_* = \operatorname{argmin}_{v\in \mathbb{R}^d} \mathbb{E}_{X\sim p_X(c),\epsilon\sim \mathcal{N}(0,1),t} \left[ \|\epsilon-\phi(X(t),t,\rho(v)) \|^2 \right] 
\end{equation}

The beauty of this framework lies in the fact that the text encoder and the diffusion model remain frozen or unchanged, while a single textual token (i.e., word embedding) is optimized to enable the diffusion model to generate collages of important patches. This significantly increases the rate of dataset compression, as it allows us to represent an entire category of images with a $d$-dimensional vector, $v^c_*$, where, for instance, $d=768$. Figure \ref{fig:step2} demonstrates this process.

\begin{figure}
    \centering
    \includegraphics[width=\linewidth]{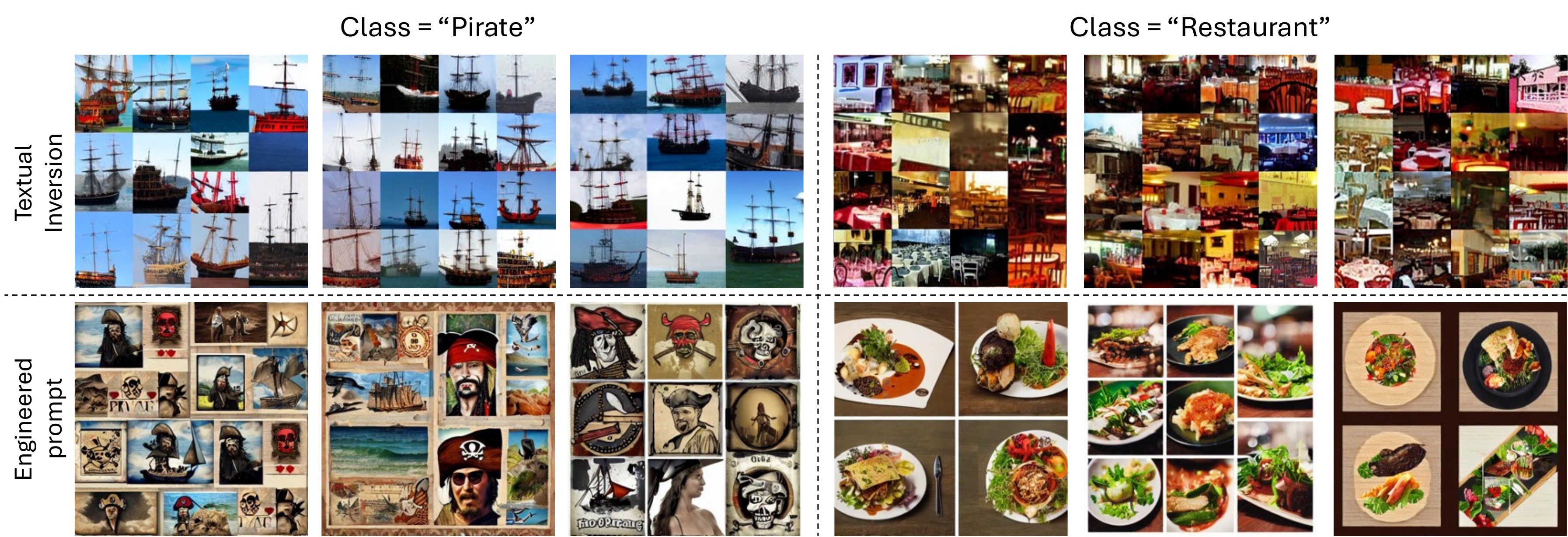}
    \caption{Comparison of collage images generated by textual inversion versus the engineered prompt, ``A 4$\times$4 natural collage of `name\_of\_class' images,'' for classes `pirate' and `restaurant' in ImageNet-1k dataset.}
    \label{fig:engineered_prompt}
\end{figure}

Note that, with abuse of notation, we denote the entire inverse diffusion process as $X^c=\Phi(\epsilon, \rho(v^c_*))$. Moreover, to demonstrate the effectiveness of the textual inversion framework for generating collage images, we provide a qualitative comparison between the generated collage images using the optimized prompt $\Phi(\epsilon, \rho(v^c_*))$, versus using an engineered prompt like ``A 4$\times$4 natural collage of `name\_of\_class' images,'' in Figure \ref{fig:engineered_prompt}. We can clearly see that the textual inversion recovers images that are closer to realistic collages both visually and semantically. The quantitative results of these experiments are included in the supplementary material. 

\begin{figure}[h!]
    \centering
    \vspace{-.2in}
    \includegraphics[width=\linewidth]{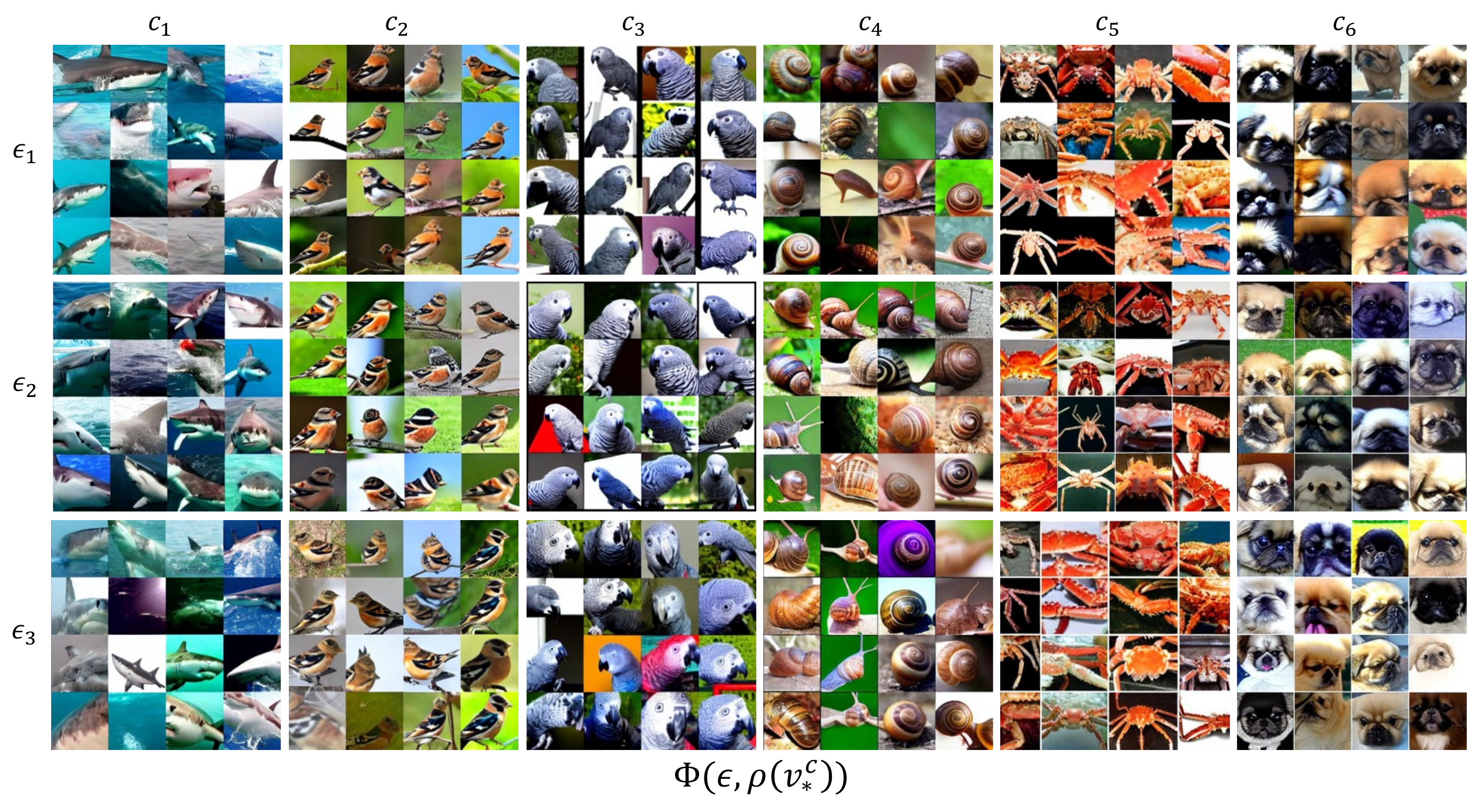}    
    \caption{Generated images via $\Phi(\epsilon, \rho(v^c_*))$ for different $c$s and different $\epsilon$s.}
    \vspace{-.2in}
    \label{fig:collages}
\end{figure}

Lastly, we show randomly generated collages, $\Phi(\epsilon, \rho(v^c_*))$, for six classes of animals and for three different random seeds in Figure \ref{fig:collages}. As can be clearly seen, the generated collages are: 1) realistic, 2) diverse, and 3) focus on discriminative characteristics of their corresponding classes, making them ideal for training a classifier.
Next, we describe the soft labeling process used in our framework.

\subsection{Step 3: One Hot vs. Soft Labeling}

For dataset distillation utilizing one-hot labels, storing $[v_*^c, c]$ is sufficient to represent category $c$. However, recent studies \cite{yin2024squeeze,sun2023diversity} highlight the advantages of employing soft-labeling techniques, leveraging a pre-trained teacher model. To incorporate soft labeling into our framework, we initially recognize that for a fixed $\rho(v)$, the stochasticity of the diffusion model can be encapsulated within the random generator seed. This implies that possessing the random generator seed alongside $\rho(v)$ uniquely identifies a collage image. Given a fixed seed, after generating a collage image for category $c$ via $\Phi(\epsilon, \rho(v_*^c))$, we partition it into its constituent patches and feed these patches through the pre-trained and frozen teacher model to derive the soft labels. Additionally, alongside $[v_*^c, c]$, we store the random generator seed and the computed soft labels to facilitate training the classifier with these images and soft labels. Notably, for higher IPCs, it suffices to store more seeds (a scalar per image) and the corresponding soft labels, which proves to be significantly more economical than preserving entire images.


\begin{figure}[t!]
    \centering
    \includegraphics[width=\linewidth]{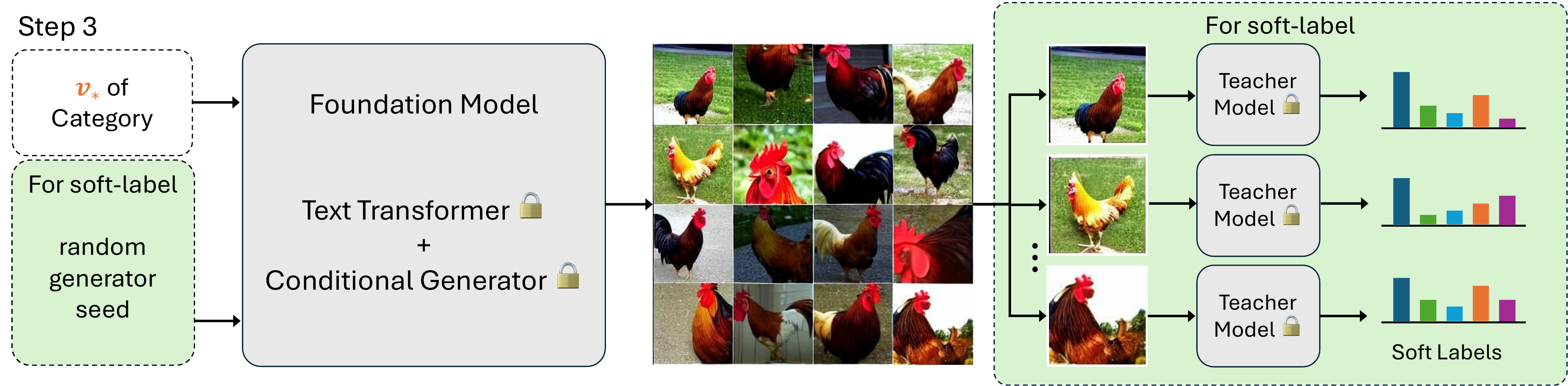}
    \caption{We conduct dataset distillation under two distinct settings. In the one-hot setting, each image category is solely represented by the prompt, $v_*$, derived from the textual inversion process illustrated in Figure \ref{fig:step2}. Conversely, in the soft-label setting, alongside the prompt $v_*$, we retain the random generator seed and the soft labels corresponding to the generated patches. It's noteworthy that the soft labels and seed will scale linearly with IPC, while the prompt remains fixed for the entire category, thus remaining independent of IPC.}
    \vspace{-.2in}
    \label{fig:step3}
\end{figure}

\section{Experiments}

To demonstrate the effectiveness of D3M, we performed extensive experiments on several large-scale and small-scale benchmarks. Below we describe the details of our experiments.

\subsection{Datasets} 
Similar to the training protocol in the prior works in the literature \cite{cazenavette2023generalizing,yin2024squeeze,sun2023diversity}, we used the following benchmark datasets:

\begin{enumerate}
    \item \textbf{CIFAR-10 \cite{krizhevsky2009learning}:} CIFAR-10 is consisted of 10 classes with images having $32 \times 32$ resolution. Similar to previous works, we used the a 128-width, 3-layer CNN (ConvNet-3) for distilling and evaluating the synthetic images. 

    \item \textbf{CIFAR-100 \cite{krizhevsky2009learning}:} CIFAR-100 contains 100 classes with the spatial resolution of $32 \times 32$. We used a ConvNet-3 in this scenario as well. 

    \item \textbf{Tiny-Imagenet \cite{le2015tiny}:} Tiny-ImageNet is a 200-class subset of ImageNet with $64 \times 64$ images. For the Tiny-ImageNet in small-scale experiments, we used the ConvNet-4, and for its large-scale version we utilized a ResNet-18. \cite{he2016deep}. 

    \item \textbf{ImageNet-100 \cite{russakovsky2015imagenet}:} ImageNet-100 consists of 100 classes of ImageNet-1k with the resolution of $224 \times 224$. ResNet-18 has been used both for the distillation and evaluation. 

    \item \textbf{ImageNet-1k \cite{deng2009imagenet}:} We used the ImageNet-1k with 1000 classes of natural images and standard resolution of $224 \times 224$. Similar to the previous large-scale experiments, we employed a ResNet-18. 
\end{enumerate}

\subsection{Baselines}
 We divided the methods into two main categories: 1) Bi-level-optimization-based and 2) Knowledge-distillation-based methods.  Here, we briefly introduce our selected baselines and their categories. For bilevel-optimization approaches, we used the following baselines:
\begin{itemize}
    \item \textbf{MTT} \cite{cazenavette2022dataset} proposes to first generate a dataset of expert trajectories and then poses a bilevel optimization problem to match the trajectory of the student to those of the teachers. 

    \item \textbf{IDM} \cite{zhao2023improved} proposes to match the output distribution of the synthetic and the real images while improving the computational cost of the previous works. 

    \item \textbf{TESLA} \cite{cui2023scaling} solves a very similar objective to MTT's \cite{cazenavette2022dataset} by calculating the exact unrolled gradients with a constant  memory complexity. 

    \item \textbf{DATM} \cite{guo2023towards} distinguishes the early and late stages of the expert trajectories and proposes a difficulty-aware solution for the trajectory matching. 
\end{itemize}

\noindent For the knowledge-distillation-based approaches we used these recent baselines:
\begin{itemize}
    \item \textbf{SRe2L} \cite{yin2024squeeze} is the first knowledge-distillation-based dataset distillation or condensation method. They propose to synthesize the distilled images using ideas from model inversion and, more particularly, utilizing the batch norm statistics of the teacher network. 

    \item \textbf{CDA} \cite{yin2023dataset} is another knowledge-distillation-based method that proposes a curriculum for the parameters of RandomResizedCrop augmentation while synthesizing the data. 

    \item \textbf{RDED} \cite{sun2023diversity} is a nascent method that introduces collages as an effective technique for dataset condensation. They showed that the important patches within each collage help with incorporating the pixel budget more efficiently.
\end{itemize}

\textbf{Experimental setup:} We evaluated the knowledge-distillation-based methods as well as ours in the following two scenarios: 1) utilizing one-hot labels and performing augmentations on the distilled data, 2) storing the soft labels per image but without augmentations. 
We emphasize that in the learning-with-soft-labels setting, one cannot utilize data augmentation. Doing so would render the stored soft labels obsolete and would require access to the teacher model to update the soft labels, which defeats the purpose of dataset distillation. 
We provide a fair comparison of the baselines under the specified settings. We repeat each experiment three times and report the mean and standard deviation for three different image-per-category (IPC) values, namely IPC $\in \{1, 10, 50\}$.

The large-scale and small-scale results are summarized in Tables \ref{tab:tab_large_scale} and \ref{tab:tab_small_scale}, respectively. Importantly, some bilevel-optimization-based methods, such as \cite{cui2023scaling}, also learn the soft labels along with the synthetic data. We evaluated these approaches as suggested in their respective papers. Please note that each collage will be cut and resized into the allowed IPC budget, following the approach outlined in \cite{sun2023diversity}. Specifically, in CIFAR-10, CIFAR-100, and Tiny-ImageNet, the collage used for training consists of just one patch of the class instance, resized to dimensions of $32 \times 32$, $32 \times 32$, and $64 \times 64$, respectively. For ImageNet-100 and ImageNet-1K, 2$\times$2 collage images are resized to a spatial resolution of $224 \times 224$, containing four patches in total. Lastly, we investigate the effect of the number of patches in the collage images on the classification accuracy of the trained classifier in our ablation studies.

\vspace{.2in}

\begin{table}[h]
\centering
\begin{adjustbox}{width=\textwidth,center}
{\renewcommand{\arraystretch}{1.3}
\begin{tabular}{c|c|c|c|c|c|c|c|c|c|c|}
\cline{3-11}
\multicolumn{2}{c|}{} & \multicolumn{3}{c|}{Tiny ImageNet} & \multicolumn{3}{c|}{ImageNet 100} & \multicolumn{3}{c|}{ImageNet 1k} \\ \cline{2-11}
\multirow{2}{*}{} & Method  & IPC=1 & IPC=10 & IPC=50 & IPC=1 & IPC=10 & IPC=50 & IPC=1 & IPC=10 & IPC=50 \\ 
\hline
\multirow{4}{*}{\rotatebox[origin=c]{90}{Hot Labels}} & SRe2L  & 1.64(.06) & 4.04(.06) & 18.39(.49) & 1.75(.02) & 4.34(.18) & 10.31(.39) & 0.26(.03) & 1.81(.06) & 4.09(.09) \\
 & CDA  & 1.16(.03) & 2.74(.03) & 14.98(.37) & 1.53(.02) & 4.13(.25) & 10.33(.53) & 0.22(.02) & 1.48(.01) & 5.83(.14) \\
 & RDED  & 3.00(.07) & 9.99(.25) & 24.59(.12) & 5.23(.10) & 14.64(.73) & 35.91(.41) & 1.12(.03) & 8.61(.10) & 26.28(.15) \\
 & Ours  & 4.98(.06) & 14.16(.64) & 18.46 (.65) & 7.57(.24) & 16.42(1.41) & 16.96(1.08) & 6.22(.10) & 12.27 (.57) & 12.38 (.67) \\ \hline
\multirow{4}{*}{\rotatebox[origin=c]{90}{Soft Labels}} & SRe2L &  2.36(.06) & 9.50(.30) & 27.1(.29) & 2.41(.12) & 6.85(.31) & 16.42(.81) & 0.44(.02) & 1.45(.03) & 5.56(.10) \\
 & CDA  & 2.42(.19) & 9.36(.27) & 26.40(.41) & 1.80(.31) & 7.22(.01) & 15.56(.55) & 0.51(.01) & 1.44(.05) &  5.83(.14) \\
 & RDED & 5.06(.18) & 19.67(.45) & 41.45(.07) & 6.85(.50) & 24.13(.10) & 49.51(.18) & 1.66(.04) & 6.40(.13) & 19.60(.10) \\
& Ours & 11.36(.51) & 38.76(.30) & 51.43 (.24) & 13.1(1.02) & 42.27(.54) & 45.25(.43) & 5.04 (.15) & 23.57(.05) & 32.23 (.12) \\
\hline
\end{tabular}}
\end{adjustbox}
\vspace{.1in}
\caption{Accuracy of knowledge-based distillation approaches in large-scale experiments on Tiny ImageNet, ImageNet 100, and ImageNet 1K, with IPC values of 1, 10, and 50. Results are presented for both one-hot labels and soft labels settings. The experiments are repeated three times, and the means and standard deviations (stds) are reported, with stds shown in parentheses. For these experiments, we used a ResNet-18 both for generating the soft labels and for the evaluation stage.}
\label{tab:tab_large_scale}
\end{table}

\begin{table}[t]
\centering
\begin{adjustbox}{width=\textwidth,center}{\renewcommand{\arraystretch}{1.3}
\begin{tabular}{c|c|c|c|c|c|c|c|c|c|c|}
\cline{3-11}
\multicolumn{2}{c|}{} & \multicolumn{3}{c|}{CIFAR-10} & \multicolumn{3}{c|}{CIFAR-100} & \multicolumn{3}{c|}{Tiny ImageNet} \\ \cline{2-11}
\multirow{2}{*}{} & Method& IPC=1 & IPC=10 & IPC=50 & IPC=1 & IPC=10 & IPC=50 & IPC=1 & IPC=10 & IPC=50 \\ 
\hline
\multirow{4}{*}{\rotatebox[origin=c]{90}{BiLevel Opt.}} & MTT & 46.3(.8) & 65.3(.7) & 71.6(.3) & 24.3(.3) & 40.1(.4) & 47.7(.2) & 8.8(.3) & 23.2(.2) & 28.0(.3) \\
 & IDM & 45.6(.7) & 58.6(.1) & 67.5(.1) & 20.1(.3) & 45.1(.1) & 50.0(.2) & 10.1(.2) & 21.9(.3) & 27.7(.3) \\
 & TESLA & 48.5(.8) & 66.4(.8) & 72.6(.7) & 24.8(.5) & 41.7(.3) & 47.9(.3) & - & - & - \\
 & DATM & 46.9(.5) & 66.8(.3) & 76.1(.3) & 27.9(.2) & 47.2(.4) & 55.0(.2) & 17.1(.3) & 31.1(.3) & 39.7(.3) \\ \hline
\multirow{4}{*}{\rotatebox[origin=c]{90}{Hot Labels}} & SRe2L  & 14.12(.97) & 20.51(.45) & 32.03(.31) & 3.11(.08) & 7.44(.21) & 13.81(.17) & 1.64(.11) & 7.23(.13) & 13.62(.30) \\
 & CDA  & 14.71(.68) & 22.88(.08) & 32.41(.35) & 3.24(.06) & 7.52(.21) & 14.07(.41) & 2.23(.07) & 4.00 (.09) & 6.03(.04) \\
 & RDED  & 20.05(.14) & 31.78(.12) & 46.53(.18) & 5.21(.10) & 12.71(.06) & 28.35(.69) & 1.88(.02) & 6.08(.11) & 16.12(.14)  \\
 & Ours  & 28.43(.09) & 37.69(.49) & 47.09(.30) & 9.26(.25) & 22.98(.13) & 29.06(.23) & 4.44(.19) & 10.80(.28) & 14.03(.16)  \\ \hline
\multirow{4}{*}{\rotatebox[origin=c]{90}{Soft Labels}} & SRe2L & 14.23(.30) & 26.07(.31) & 37.26(.92) & 5.08(.06) & 24.54(.09) & 39.24(.38) & 2.14(.07) & 10.29(.34) & 25.10(.12) \\
 & CDA  & 16.21(.28) & 25.64(.59) & 36.24(.52) & 5.10(.07) & 24.15(.44) & 38.99(.18) & 2.56(.11) & 7.66(.19) & 15.98(.48) \\
 & RDED & 15.50(.66) & 38.64(1.29) & 56.28(.33) & 10.77(.16) & 35.68(.42) & 47.29(.10) & 4.68(.15) & 21.11(.40) & 37.81(.31) \\
& Ours & 35.88(.07) & 58.58(.12) & 70.52(.29) & 30.78(.37) & 49.09(.04) & 54.51(.04) & 12.87(.05) & 37.60(.25) & 47.77(.13) \\
\hline
\end{tabular}}
\end{adjustbox}
\vspace{.1in}
\caption{Performance of various dataset condensation techniques on small-scale benchmarks. Standard ConvNet-3 architecture was employed for CIFAR-10 and CIFAR-100, while the ConvNet-4 architecture was used for Tiny-ImageNet, following prior literature. TESLA \cite{cui2023scaling} did not report accuracies for Tiny-ImageNet, and consistent with previous studies \cite{sun2023diversity, guo2023towards}, its entries are left blank.}
\label{tab:tab_small_scale}
\end{table}

\vspace{-.3in}
\subsection{Accuracy vs. Compression}
While Tables \ref{tab:tab_large_scale} and \ref{tab:tab_small_scale} showcase the remarkable performance of our proposed single-prompt dataset distillation framework, D3M, they do not entirely capture the compression benefits inherent in the proposed method. It is crucial to highlight that higher compression rates would translate to more efficient communication and storage, particularly in bandwidth-constrained environments. In this context, we present the accuracy of various dataset distillation methods relative to the size of compressed data utilized for training the classifier network.

\begin{figure}[t!]
    \centering
    \includegraphics[width=\linewidth]{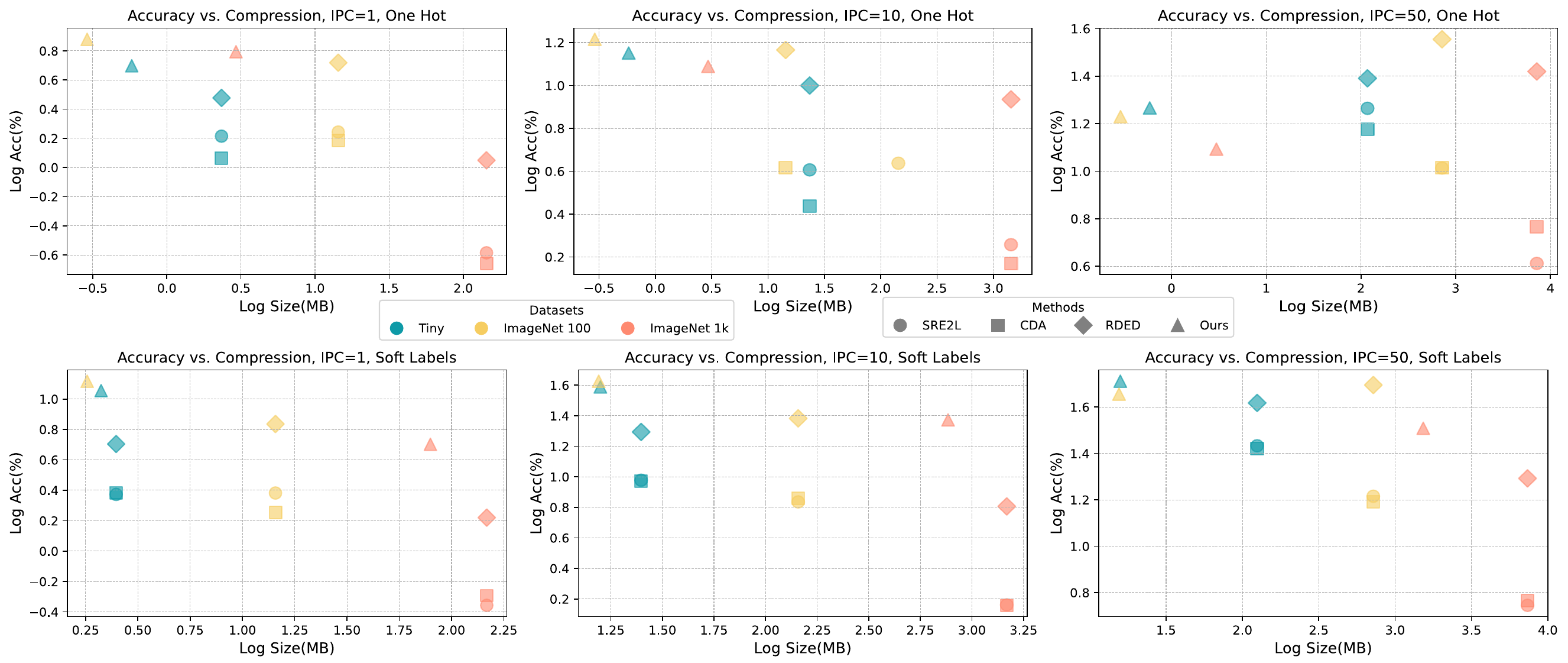}
    \caption{ Log-log plot illustrating the performance of dataset distillation methods on large-scale datasets, including Tiny ImageNet, ImageNet 100, and ImageNet 1K, across IPC values of 1, 10, and 50. The performance is depicted as a function of the size of compressed data utilized for training the classifier network. Methods closer to the top left demonstrate superior performance.}
    \label{fig:storage}
\end{figure}

\begin{table}[t]
\begin{adjustbox}{width=.65\textwidth,center}{\renewcommand{\arraystretch}{1.3}}
\begin{tabular}{cc|c|c|c|}
\cline{3-5}
& & {\footnotesize ResNet-18}  & {\footnotesize MobileNet-v2} & {\footnotesize DenseNet-121} \\
\midrule
\multicolumn{1}{|c|}{SRe2L} & ResNet-18  & 1.45 & 0.84 & 5.60 \\
\midrule
\multicolumn{1}{|c|}{ } & ResNet-18 (11.5M)  & 12.81 & 9.59 & 20.14 \\
\multicolumn{1}{|c|}{Ours} & MobileNet-v2 (3.4M)& 12.12 & 9.66 & 19.45 \\
\multicolumn{1}{|c|}{ } & DenseNet-121 (7.97M) & 10.64 & 10.31 & 17.24 \\
\bottomrule
\end{tabular}
\end{adjustbox}
\vspace{.1in}
\caption{Cross-architecture evaluation of the synthetic data across ResNet-18, MobileNet-v2, and DenseNet-121. }
\label{tab:cross_arch}
\end{table}

We observe that an advantage of storing the prompts as a distributional representation of the data, rather than the individual samples, is the flexibility to introduce variations to the data within a fixed running-memory and computation budget, with minimal overhead of only storing the random seed and patch-level soft labels or collage-level one-hot labels. Consequently, in our experiments, these new variations of the data replace the old ones without consuming additional memory, aside from potential extra soft labels. Moreover, the total number of iterations remains constant to ensure a fair comparison. Figure \ref{fig:storage} illustrates the performance of dataset distillation methods for the three IPCs as a function of the size of compressed data used for training the classifier network. We see that D3M exhibits high efficiency in compressing the datasets, particularly for higher IPCs, in large-scale experiments while maintaining accuracy.

\subsection{Cross-architecture analysis}
In order to demonstrate the generalizability of our synthetic data across various architectures, we performed a cross-architectural analysis while fixing the $IPC=10$ in ImageNet-1k. We first generated our collages using the learned prompts. Then, we assigned soft labels to the synthetic data using various pre-trained models on ImageNet-1k. We used the ResNet-18, MobileNet-v2, and DenseNet-121 in our experiments. In the final stage, we trained several student models from scratch on the synthetic images and their corresponding soft labels. The results are cross-examined and shown in Table \ref{tab:cross_arch}. Moreover, in order to better understand the generalizability power of our distilled data, unlike \cite{sun2023diversity}, we do not perform data augmentation nor replace our data with their variations during the training stage. Table \ref{tab:cross_arch} shows that depending on the different teacher/student pairs, using a different model for the evaluation can be an effective strategy. For instance, going from ResNet-18 with 11.5 million parameters to DenseNet-121 with 8 million parameters can lead to performance improvement while reducing the model size. We speculate that by synthesizing images close to the distribution of natural images, we can significantly reduce the architecture-specific biases in the synthesis process. 

\begin{figure}[t!]
  \centering
    \includegraphics[width=.75\linewidth]{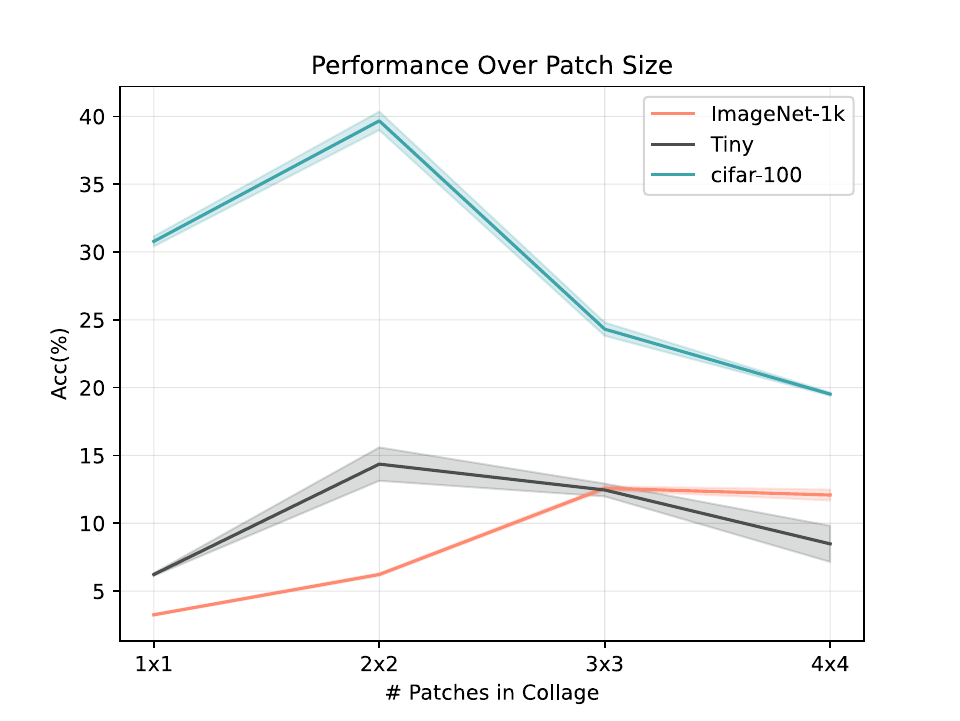}
    \caption{Accuracy over different collage patch sizes, across various datasets. The results demonstrate that there is an optimal value of patch sizes depending on the original resolution of the dataset. Nevertheless, for high patch sizes, the performance deteriorates due to the loss of information in the squeezed images.}
    \label{fig:patch_size}
\end{figure}


\subsection{Ablation: Patch Size}

In this ablation study, we studied the effect of different patch sizes in the generated collages. We experimented with $1 \times 1 $, $2 \times 2 $, $3 \times 3 $, and $4 \times 4 $ patches in $IPC=1$ setting of ImageNet-1k. Here, we report our results in the soft-label, no augmentation and variation scenario. The results are shown in Figure \ref{fig:patch_size}. We can observe that for high patch sizes, performance diminishes, especially in the low-resolution setting of CIFAR-100. This is due to the fact that too much details are squeezed inside high patch sizes of a collage and we will loose the information of the important patches. Please note that although increasing the patch sizes beyond $2 \times 2$ in ImageNet-1k leads to better performance, the memory requirement for storing the soft labels will be more than twice as much. 

\subsection{Ablation: Steps 1 and 2}
We also conducted additional ablation studies to illustrate the necessity of steps 1 and 2 in our framework. For step 1, instead of selecting the important patches, we simply resized and center-cropped the input images to construct patches. This alternative approach exhibited inferior performance compared to selecting the important patches when training the classifier. Regarding step 2, alongside the qualitative results presented in Section 3.3, we quantitatively demonstrated that merely generating training images using the class label and an engineered prompt for the diffusion model resulted in a significant drop in performance compared to textual inversion. These findings are detailed in our supplementary materials due to space constraints.

\section{Conclusion}

This paper originates from a scientific curiosity about the compressibility of large-scale image datasets when given access to a foundational text-to-image model. To explore this, we introduced Dataset Distillation using Diffusion Models (D3M), which builds upon the latest advances in large-scale dataset distillation methods. D3M utilizes collage images of important patches in conjunction with textual inversion and text-to-image diffusion models, achieving unprecedented compression of ImageNet-scale datasets. We demonstrate that D3M can condense an entire category of images into a single textual prompt, illustrating its powerful condensation capability. Through comprehensive experiments conducted as part of this study, we illustrate that D3M can achieve superior dataset compression rates while still resulting in high performance when training a classifier on the condensed data. This characteristic is mainly due to the realistic and diverse nature of the generated images from the prompted diffusion model.

\bibliographystyle{unsrtnat}
\bibliography{main_arxiv}  

\begin{thebibliography}{47}
\providecommand{\natexlab}[1]{#1}
\providecommand{\url}[1]{\texttt{#1}}
\expandafter\ifx\csname urlstyle\endcsname\relax
  \providecommand{\doi}[1]{doi: #1}\else
  \providecommand{\doi}{doi: \begingroup \urlstyle{rm}\Url}\fi

\bibitem[Krasin et~al.(2017)Krasin, Duerig, Alldrin, Ferrari, Abu-El-Haija, Kuznetsova, Rom, Uijlings, Popov, Kamali, Malloci, Pont-Tuset, Veit, Belongie, Gomes, Gupta, Sun, Chechik, Cai, Feng, Narayanan, and Murphy]{OpenImages2}
Ivan Krasin, Tom Duerig, Neil Alldrin, Vittorio Ferrari, Sami Abu-El-Haija, Alina Kuznetsova, Hassan Rom, Jasper Uijlings, Stefan Popov, Shahab Kamali, Matteo Malloci, Jordi Pont-Tuset, Andreas Veit, Serge Belongie, Victor Gomes, Abhinav Gupta, Chen Sun, Gal Chechik, David Cai, Zheyun Feng, Dhyanesh Narayanan, and Kevin Murphy.
\newblock Openimages: A public dataset for large-scale multi-label and multi-class image classification.
\newblock \emph{Dataset available from https://storage.googleapis.com/openimages/web/index.html}, 2017.

\bibitem[Maclaurin et~al.(2015)Maclaurin, Duvenaud, and Adams]{maclaurin2015gradient}
Dougal Maclaurin, David Duvenaud, and Ryan Adams.
\newblock Gradient-based hyperparameter optimization through reversible learning.
\newblock In \emph{International conference on machine learning}, pages 2113--2122. PMLR, 2015.

\bibitem[Lorraine et~al.(2020)Lorraine, Vicol, and Duvenaud]{lorraine2020optimizing}
Jonathan Lorraine, Paul Vicol, and David Duvenaud.
\newblock Optimizing millions of hyperparameters by implicit differentiation.
\newblock In \emph{International conference on artificial intelligence and statistics}, pages 1540--1552. PMLR, 2020.

\bibitem[Wang et~al.(2018)Wang, Zhu, Torralba, and Efros]{wang2018dataset}
Tongzhou Wang, Jun-Yan Zhu, Antonio Torralba, and Alexei~A Efros.
\newblock Dataset distillation.
\newblock \emph{arXiv preprint arXiv:1811.10959}, 2018.

\bibitem[Zhao et~al.(2020)Zhao, Mopuri, and Bilen]{zhao2020dataset}
Bo~Zhao, Konda~Reddy Mopuri, and Hakan Bilen.
\newblock Dataset condensation with gradient matching.
\newblock In \emph{International Conference on Learning Representations}, 2020.

\bibitem[Liu et~al.(2023)Liu, Ye, Yu, and Wang]{liu2023slimmable}
Songhua Liu, Jingwen Ye, Runpeng Yu, and Xinchao Wang.
\newblock Slimmable dataset condensation.
\newblock In \emph{Proceedings of the IEEE/CVF Conference on Computer Vision and Pattern Recognition}, pages 3759--3768, 2023.

\bibitem[Cazenavette et~al.(2023)Cazenavette, Wang, Torralba, Efros, and Zhu]{cazenavette2023generalizing}
George Cazenavette, Tongzhou Wang, Antonio Torralba, Alexei~A Efros, and Jun-Yan Zhu.
\newblock Generalizing dataset distillation via deep generative prior.
\newblock In \emph{Proceedings of the IEEE/CVF Conference on Computer Vision and Pattern Recognition}, pages 3739--3748, 2023.

\bibitem[Cui et~al.(2023)Cui, Wang, Si, and Hsieh]{cui2023scaling}
Justin Cui, Ruochen Wang, Si~Si, and Cho-Jui Hsieh.
\newblock Scaling up dataset distillation to imagenet-1k with constant memory.
\newblock In \emph{International Conference on Machine Learning}, pages 6565--6590. PMLR, 2023.

\bibitem[Sun et~al.(2023)Sun, Shi, Yu, and Lin]{sun2023diversity}
Peng Sun, Bei Shi, Daiwei Yu, and Tao Lin.
\newblock On the diversity and realism of distilled dataset: An efficient dataset distillation paradigm.
\newblock \emph{arXiv preprint arXiv:2312.03526}, 2023.

\bibitem[Guo et~al.(2023)Guo, Wang, Cazenavette, Li, Zhang, and You]{guo2023towards}
Ziyao Guo, Kai Wang, George Cazenavette, Hui Li, Kaipeng Zhang, and Yang You.
\newblock Towards lossless dataset distillation via difficulty-aligned trajectory matching.
\newblock \emph{arXiv preprint arXiv:2310.05773}, 2023.

\bibitem[Shao et~al.(2023)Shao, Yin, Zhou, Zhang, and Shen]{shao2023generalized}
Shitong Shao, Zeyuan Yin, Muxin Zhou, Xindong Zhang, and Zhiqiang Shen.
\newblock Generalized large-scale data condensation via various backbone and statistical matching.
\newblock \emph{arXiv preprint arXiv:2311.17950}, 2023.

\bibitem[Zhou et~al.(2023)Zhou, Wang, Gu, Peng, Lian, Zhang, You, and Feng]{zhou2023dataset}
Daquan Zhou, Kai Wang, Jianyang Gu, Xiangyu Peng, Dongze Lian, Yifan Zhang, Yang You, and Jiashi Feng.
\newblock Dataset quantization.
\newblock In \emph{Proceedings of the IEEE/CVF International Conference on Computer Vision}, pages 17205--17216, 2023.

\bibitem[Yin et~al.(2024)Yin, Xing, and Shen]{yin2024squeeze}
Zeyuan Yin, Eric Xing, and Zhiqiang Shen.
\newblock Squeeze, recover and relabel: Dataset condensation at imagenet scale from a new perspective.
\newblock \emph{Advances in Neural Information Processing Systems}, 36, 2024.

\bibitem[Chen et~al.(2024)Chen, Yang, Wang, and Mirzasoleiman]{chen2024data}
Xuxi Chen, Yu~Yang, Zhangyang Wang, and Baharan Mirzasoleiman.
\newblock Data distillation can be like vodka: Distilling more times for better quality.
\newblock In \emph{The Twelfth International Conference on Learning Representations}, 2024.
\newblock URL \url{https://openreview.net/forum?id=1NHgmKqOzZ}.

\bibitem[Cazenavette et~al.(2022)Cazenavette, Wang, Torralba, Efros, and Zhu]{cazenavette2022dataset}
George Cazenavette, Tongzhou Wang, Antonio Torralba, Alexei~A Efros, and Jun-Yan Zhu.
\newblock Dataset distillation by matching training trajectories.
\newblock In \emph{Proceedings of the IEEE/CVF Conference on Computer Vision and Pattern Recognition}, pages 4750--4759, 2022.

\bibitem[Deng et~al.(2009)Deng, Dong, Socher, Li, Li, and Fei-Fei]{deng2009imagenet}
Jia Deng, Wei Dong, Richard Socher, Li-Jia Li, Kai Li, and Li~Fei-Fei.
\newblock Imagenet: A large-scale hierarchical image database.
\newblock In \emph{2009 IEEE conference on computer vision and pattern recognition}, pages 248--255. Ieee, 2009.

\bibitem[Gal et~al.(2022)Gal, Alaluf, Atzmon, Patashnik, Bermano, Chechik, and Cohen-or]{gal2022image}
Rinon Gal, Yuval Alaluf, Yuval Atzmon, Or~Patashnik, Amit~Haim Bermano, Gal Chechik, and Daniel Cohen-or.
\newblock An image is worth one word: Personalizing text-to-image generation using textual inversion.
\newblock In \emph{The Eleventh International Conference on Learning Representations}, 2022.

\bibitem[Rombach et~al.(2022)Rombach, Blattmann, Lorenz, Esser, and Ommer]{rombach2022high}
Robin Rombach, Andreas Blattmann, Dominik Lorenz, Patrick Esser, and Bj{\"o}rn Ommer.
\newblock High-resolution image synthesis with latent diffusion models.
\newblock In \emph{Proceedings of the IEEE/CVF conference on computer vision and pattern recognition}, pages 10684--10695, 2022.

\bibitem[Brown et~al.(2020)Brown, Mann, Ryder, Subbiah, Kaplan, Dhariwal, Neelakantan, Shyam, Sastry, Askell, et~al.]{brown2020language}
Tom Brown, Benjamin Mann, Nick Ryder, Melanie Subbiah, Jared~D Kaplan, Prafulla Dhariwal, Arvind Neelakantan, Pranav Shyam, Girish Sastry, Amanda Askell, et~al.
\newblock Language models are few-shot learners.
\newblock \emph{Advances in neural information processing systems}, 33:\penalty0 1877--1901, 2020.

\bibitem[Radford et~al.(2021)Radford, Kim, Hallacy, Ramesh, Goh, Agarwal, Sastry, Askell, Mishkin, Clark, et~al.]{radford2021learning}
Alec Radford, Jong~Wook Kim, Chris Hallacy, Aditya Ramesh, Gabriel Goh, Sandhini Agarwal, Girish Sastry, Amanda Askell, Pamela Mishkin, Jack Clark, et~al.
\newblock Learning transferable visual models from natural language supervision.
\newblock In \emph{International conference on machine learning}, pages 8748--8763. PMLR, 2021.

\bibitem[Kim et~al.(2022)Kim, Kim, Oh, Yun, Song, Jeong, Ha, and Song]{kim2022dataset}
Jang-Hyun Kim, Jinuk Kim, Seong~Joon Oh, Sangdoo Yun, Hwanjun Song, Joonhyun Jeong, Jung-Woo Ha, and Hyun~Oh Song.
\newblock Dataset condensation via efficient synthetic-data parameterization.
\newblock In \emph{International Conference on Machine Learning}, pages 11102--11118. PMLR, 2022.

\bibitem[Zhang et~al.(2023)Zhang, Zhang, Lei, Mukherjee, Pan, Zhao, Ding, Li, and Xu]{zhang2023accelerating}
Lei Zhang, Jie Zhang, Bowen Lei, Subhabrata Mukherjee, Xiang Pan, Bo~Zhao, Caiwen Ding, Yao Li, and Dongkuan Xu.
\newblock Accelerating dataset distillation via model augmentation.
\newblock In \emph{Proceedings of the IEEE/CVF Conference on Computer Vision and Pattern Recognition}, pages 11950--11959, 2023.

\bibitem[Wang et~al.(2022)Wang, Zhao, Peng, Zhu, Yang, Wang, Huang, Bilen, Wang, and You]{wang2022cafe}
Kai Wang, Bo~Zhao, Xiangyu Peng, Zheng Zhu, Shuo Yang, Shuo Wang, Guan Huang, Hakan Bilen, Xinchao Wang, and Yang You.
\newblock Cafe: Learning to condense dataset by aligning features.
\newblock In \emph{Proceedings of the IEEE/CVF Conference on Computer Vision and Pattern Recognition}, pages 12196--12205, 2022.

\bibitem[Zhao and Bilen(2023)]{zhao2023dataset}
Bo~Zhao and Hakan Bilen.
\newblock Dataset condensation with distribution matching.
\newblock In \emph{Proceedings of the IEEE/CVF Winter Conference on Applications of Computer Vision}, pages 6514--6523, 2023.

\bibitem[Cui et~al.(2022)Cui, Wang, Si, and Hsieh]{cui2022dc}
Justin Cui, Ruochen Wang, Si~Si, and Cho-Jui Hsieh.
\newblock Dc-bench: Dataset condensation benchmark.
\newblock \emph{Advances in Neural Information Processing Systems}, 35:\penalty0 810--822, 2022.

\bibitem[Jacot et~al.(2018)Jacot, Gabriel, and Hongler]{jacot2018neural}
Arthur Jacot, Franck Gabriel, and Cl{\'e}ment Hongler.
\newblock Neural tangent kernel: Convergence and generalization in neural networks.
\newblock \emph{Advances in neural information processing systems}, 31, 2018.

\bibitem[Novak et~al.(2019)Novak, Xiao, Hron, Lee, Alemi, Sohl-Dickstein, and Schoenholz]{novak2019neural}
Roman Novak, Lechao Xiao, Jiri Hron, Jaehoon Lee, Alexander~A Alemi, Jascha Sohl-Dickstein, and Samuel~S Schoenholz.
\newblock Neural tangents: Fast and easy infinite neural networks in python.
\newblock In \emph{International Conference on Learning Representations}, 2019.

\bibitem[Nguyen et~al.(2021)Nguyen, Novak, Xiao, and Lee]{nguyen2021dataset}
Timothy Nguyen, Roman Novak, Lechao Xiao, and Jaehoon Lee.
\newblock Dataset distillation with infinitely wide convolutional networks.
\newblock In A.~Beygelzimer, Y.~Dauphin, P.~Liang, and J.~Wortman Vaughan, editors, \emph{Advances in Neural Information Processing Systems}, 2021.
\newblock URL \url{https://openreview.net/forum?id=hXWPpJedrVP}.

\bibitem[Loo et~al.(2023)Loo, Hasani, Lechner, and Rus]{loo2023dataset}
Noel Loo, Ramin Hasani, Mathias Lechner, and Daniela Rus.
\newblock Dataset distillation with convexified implicit gradients.
\newblock \emph{arXiv preprint arXiv:2302.06755}, 2023.

\bibitem[Mirzasoleiman et~al.(2020)Mirzasoleiman, Bilmes, and Leskovec]{mirzasoleiman2020coresets}
Baharan Mirzasoleiman, Jeff Bilmes, and Jure Leskovec.
\newblock Coresets for data-efficient training of machine learning models.
\newblock In \emph{International Conference on Machine Learning}, pages 6950--6960. PMLR, 2020.

\bibitem[Pooladzandi et~al.(2022)Pooladzandi, Davini, and Mirzasoleiman]{pooladzandi2022adaptive}
Omead Pooladzandi, David Davini, and Baharan Mirzasoleiman.
\newblock Adaptive second order coresets for data-efficient machine learning.
\newblock In \emph{International Conference on Machine Learning}, pages 17848--17869. PMLR, 2022.

\bibitem[Maalouf et~al.(2022)Maalouf, Eini, Mussay, Feldman, and Osadchy]{maalouf2022unified}
Alaa Maalouf, Gilad Eini, Ben Mussay, Dan Feldman, and Margarita Osadchy.
\newblock A unified approach to coreset learning.
\newblock \emph{IEEE Transactions on Neural Networks and Learning Systems}, 2022.

\bibitem[Borsos et~al.(2020)Borsos, Mutny, and Krause]{borsos2020coresets}
Zal{\'a}n Borsos, Mojmir Mutny, and Andreas Krause.
\newblock Coresets via bilevel optimization for continual learning and streaming.
\newblock \emph{Advances in neural information processing systems}, 33:\penalty0 14879--14890, 2020.

\bibitem[Toneva et~al.(2018)Toneva, Sordoni, des Combes, Trischler, Bengio, and Gordon]{toneva2018empirical}
Mariya Toneva, Alessandro Sordoni, Remi~Tachet des Combes, Adam Trischler, Yoshua Bengio, and Geoffrey~J Gordon.
\newblock An empirical study of example forgetting during deep neural network learning.
\newblock In \emph{International Conference on Learning Representations}, 2018.

\bibitem[Paul et~al.(2021)Paul, Ganguli, and Dziugaite]{paul2021deep}
Mansheej Paul, Surya Ganguli, and Gintare~Karolina Dziugaite.
\newblock Deep learning on a data diet: Finding important examples early in training.
\newblock \emph{Advances in Neural Information Processing Systems}, 34:\penalty0 20596--20607, 2021.

\bibitem[Fredrikson et~al.(2015)Fredrikson, Jha, and Ristenpart]{fredrikson2015model}
Matt Fredrikson, Somesh Jha, and Thomas Ristenpart.
\newblock Model inversion attacks that exploit confidence information and basic countermeasures.
\newblock In \emph{Proceedings of the 22nd ACM SIGSAC conference on computer and communications security}, pages 1322--1333, 2015.

\bibitem[Yin et~al.(2020)Yin, Molchanov, Alvarez, Li, Mallya, Hoiem, Jha, and Kautz]{yin2020dreaming}
Hongxu Yin, Pavlo Molchanov, Jose~M Alvarez, Zhizhong Li, Arun Mallya, Derek Hoiem, Niraj~K Jha, and Jan Kautz.
\newblock Dreaming to distill: Data-free knowledge transfer via deepinversion.
\newblock In \emph{Proceedings of the IEEE/CVF Conference on Computer Vision and Pattern Recognition}, pages 8715--8724, 2020.

\bibitem[Zhou et~al.(2016)Zhou, Khosla, Lapedriza, Oliva, and Torralba]{zhou2016learning}
Bolei Zhou, Aditya Khosla, Agata Lapedriza, Aude Oliva, and Antonio Torralba.
\newblock Learning deep features for discriminative localization.
\newblock In \emph{Proceedings of the IEEE conference on computer vision and pattern recognition}, pages 2921--2929, 2016.

\bibitem[Selvaraju et~al.(2017)Selvaraju, Cogswell, Das, Vedantam, Parikh, and Batra]{selvaraju2017grad}
Ramprasaath~R Selvaraju, Michael Cogswell, Abhishek Das, Ramakrishna Vedantam, Devi Parikh, and Dhruv Batra.
\newblock Grad-cam: Visual explanations from deep networks via gradient-based localization.
\newblock In \emph{Proceedings of the IEEE international conference on computer vision}, pages 618--626, 2017.

\bibitem[Chattopadhay et~al.(2018)Chattopadhay, Sarkar, Howlader, and Balasubramanian]{chattopadhay2018grad}
Aditya Chattopadhay, Anirban Sarkar, Prantik Howlader, and Vineeth~N Balasubramanian.
\newblock Grad-cam++: Generalized gradient-based visual explanations for deep convolutional networks.
\newblock In \emph{2018 IEEE winter conference on applications of computer vision (WACV)}, pages 839--847. IEEE, 2018.

\bibitem[Devlin et~al.(2018)Devlin, Chang, Lee, and Toutanova]{devlin2018bert}
Jacob Devlin, Ming-Wei Chang, Kenton Lee, and Kristina Toutanova.
\newblock Bert: Pre-training of deep bidirectional transformers for language understanding.
\newblock \emph{arXiv preprint arXiv:1810.04805}, 2018.

\bibitem[Krizhevsky et~al.(2009)Krizhevsky, Hinton, et~al.]{krizhevsky2009learning}
Alex Krizhevsky, Geoffrey Hinton, et~al.
\newblock Learning multiple layers of features from tiny images.
\newblock 2009.

\bibitem[Le and Yang(2015)]{le2015tiny}
Ya~Le and Xuan Yang.
\newblock Tiny imagenet visual recognition challenge.
\newblock \emph{CS 231N}, 7\penalty0 (7):\penalty0 3, 2015.

\bibitem[He et~al.(2016)He, Zhang, Ren, and Sun]{he2016deep}
Kaiming He, Xiangyu Zhang, Shaoqing Ren, and Jian Sun.
\newblock Deep residual learning for image recognition.
\newblock In \emph{Proceedings of the IEEE conference on computer vision and pattern recognition}, pages 770--778, 2016.

\bibitem[Russakovsky et~al.(2015)Russakovsky, Deng, Su, Krause, Satheesh, Ma, Huang, Karpathy, Khosla, Bernstein, et~al.]{russakovsky2015imagenet}
Olga Russakovsky, Jia Deng, Hao Su, Jonathan Krause, Sanjeev Satheesh, Sean Ma, Zhiheng Huang, Andrej Karpathy, Aditya Khosla, Michael Bernstein, et~al.
\newblock Imagenet large scale visual recognition challenge.
\newblock \emph{International journal of computer vision}, 115:\penalty0 211--252, 2015.

\bibitem[Zhao et~al.(2023)Zhao, Li, Qin, and Yu]{zhao2023improved}
Ganlong Zhao, Guanbin Li, Yipeng Qin, and Yizhou Yu.
\newblock Improved distribution matching for dataset condensation.
\newblock In \emph{Proceedings of the IEEE/CVF Conference on Computer Vision and Pattern Recognition}, pages 7856--7865, 2023.

\bibitem[Yin and Shen(2023)]{yin2023dataset}
Zeyuan Yin and Zhiqiang Shen.
\newblock Dataset distillation in large data era.
\newblock 2023.

\end{thebibliography}






\end{document}